\documentclass[letterpaper, 10 pt, conference]{ieeeconf-modified}  %

\IEEEoverridecommandlockouts                              %

\usepackage[numbers]{natbib}
\usepackage{graphicx}
\graphicspath{
    {figures/}
    {figures/learning}
    {figures/model}
    {figures/model/bolt_exammples}
    {figures/calibration}
    {figures/introduction}
    {figures/sim2real}
}
\usepackage{siunitx}
\usepackage{algorithm}
\usepackage{algorithmic}
\usepackage{amsmath} %
\usepackage{amssymb}  %
\usepackage{textcomp}
\usepackage{hyperref}
\usepackage[capitalize]{cleveref}
\usepackage{capt-of}

\DeclareMathOperator{\clip}{clip} 
\DeclareMathOperator*{\argmin}{arg\,min}
\DeclareMathOperator{\pen}{pen}
\DeclareMathOperator{\rew}{rew}
\usepackage{multirow}

\crefname{equation}{}{}

\title{\LARGE \bf
Fine Manipulation Using a Tactile Skin: \\Learning in Simulation and Sim-to-Real Transfer
}
\author{Ulf Kasolowsky and Berthold B\"auml%
\thanks{The authors are with the DLR Institute of Robotics and Mechatronics and with the Technical University of Munich (TUM). \newline 
Contact: \tt\footnotesize{ulf.kasolowsky|berthold.baeuml@tum.de}
}
}

\begin{document}
\twocolumn[{%
        \renewcommand\twocolumn[1][]{#1}%
        \maketitle
        \begin{center}
            \centering
            \vskip -0.3cm
            \includegraphics[width=0.95\textwidth]{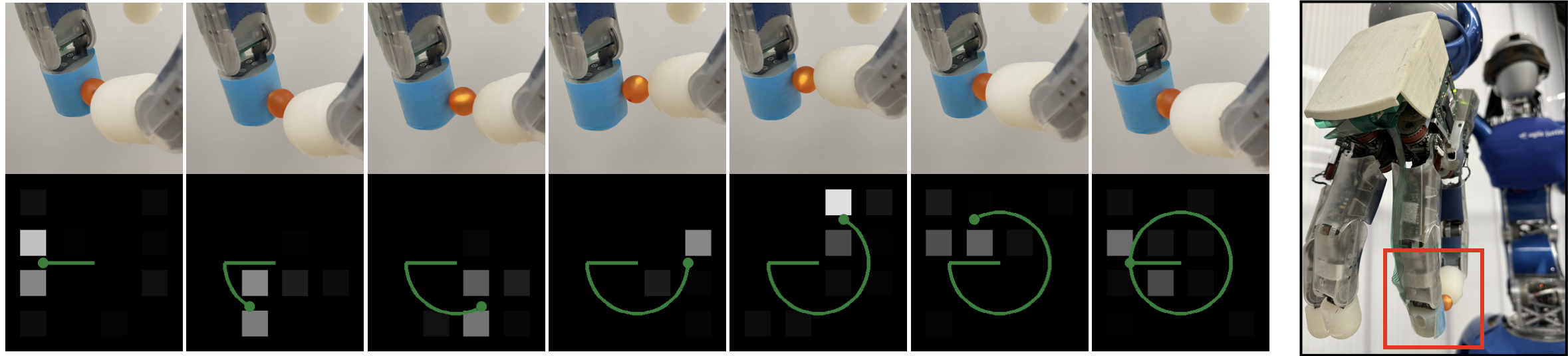}
            \vspace{-0.3cm}
            \captionof{figure}{Time sequence of a fine manipulation task ($\Delta t = \SI{1}{\second}$ between images). The marble has to be rolled between two fingers along a specified trajectory (here a circle) using the feedback from the tactile skin (blue fingertip, see also \cref{fig:tactile_sensor}, Right). The lower row shows the actual tactile image with $4\times 4$ taxels. The task is executed on the torque-controlled DLR-Hand~II~\cite{Butterfass2001} which is mounted on the humanoid robot DLR Agile Justin~\cite{Bauml.2014} (on the right). Only the hand's position and torque sensors and the tactile skin are used,  but no, e.g., visual input.}
            \label{fig:title_figure}
        \end{center}%
    }]

{\let\thefootnote\relax\footnote[0]{
   \noindent The authors are with the Learning AI for Dextrous Robots Lab \href{https://aidx-lab.org}{(aidx-lab.org)}, Technical University of Munich, Germany, and the DLR Institute of Robotics and Mechatronics, German Aerospace Center.\newline
Email:{\tt\scriptsize{ \{ulf.kasolowsky|berthold.baeuml\}@tum.de}}\newline}}
\thispagestyle{empty}
\pagestyle{empty}

\begin{abstract}

We want to enable fine manipulation with a multi-fingered robotic hand by using modern deep reinforcement learning methods. 
Key for fine manipulation is a spatially resolved tactile sensor.
Here, we present a novel model of a tactile skin that can be used together with rigid-body (hence fast) physics simulators. 
The model considers the softness of the real fingertips such that a contact can spread across multiple taxels of the sensor depending on the contact geometry. 
We calibrate the model parameters to allow for an accurate simulation of the real-world sensor. 
For this, we present a self-contained calibration method without external tools or sensors.
To demonstrate the validity of our approach, we learn two challenging fine manipulation tasks: Rolling a marble and a bolt between two fingers. 
We show in simulation experiments that tactile feedback is crucial for precise manipulation and reaching sub-taxel resolution of $<$\SI{1}{\milli\meter} (despite a taxel spacing of \SI{4}{\milli\meter}). 
Moreover, we demonstrate that all policies successfully transfer from the simulation to the real robotic hand.\\
Website: \href{https://aidx-lab.org/skin/iros24}{aidx-lab.org/skin/iros24}
\end{abstract}

\section{Introduction}
The dextrous manipulation of small objects is a key skill required for all kinds of industrial tasks like sorting or assembling.
Humans excel in fine manipulation using their dextrous hands in combination with their spatially resolved sense of touch.
In robotics, only recently, the advent of modern learning methods based on excessive training in simulation enabled breakthroughs in dextrous manipulation with multi-fingered hands, e.g., in-hand manipulation.
However, for learning fine manipulation tasks, in addition a precise but computationally efficient simulation of tactile sensors is needed. 

\begin{figure}[!htb]
    \centering
    \includegraphics[width=1.0\linewidth]{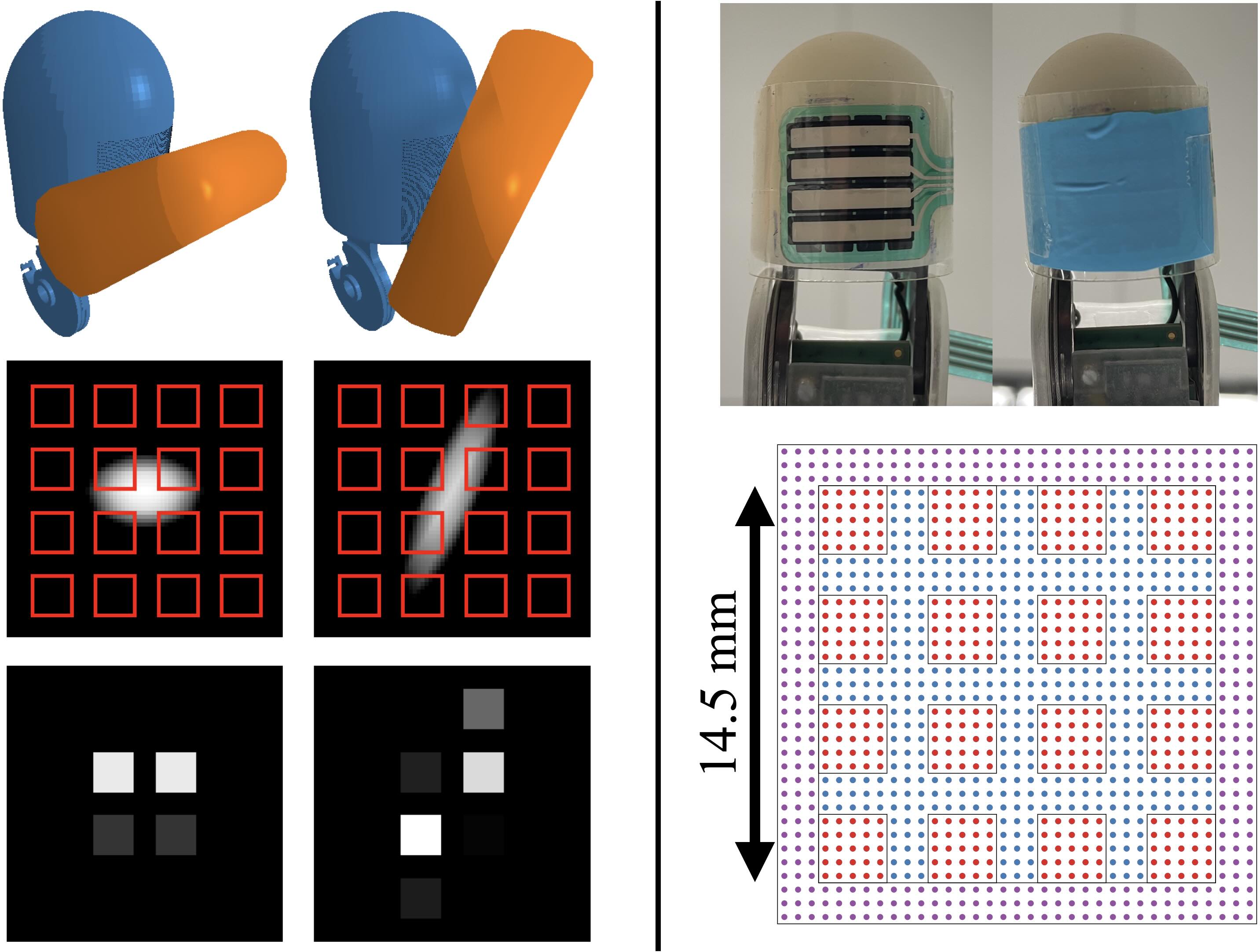}
    \vspace{-0.6cm}
    \caption{
    \textbf{Left:} Tactile response due to soft contact. The top row shows two contact situations of a bolt with a tactile sensor on a cylindrical fingertip. Only if the contact is soft, the bolt will penetrate the fingertip resulting in a contact area causing multi-taxel response instead of a single contact point. The second row depicts the pressure distribution on the fingertip and the bottom row the resulting response for the individual taxel.\\\textbf{Right:} Tactile sensor. The top row shows the sensor without and with the rubber glove covering (for better grip). The sensor has $4\times 4$ taxels, which each measure $\SI{2.5}{\milli\meter} \times \SI{2.5}{\milli\meter}$ and are arranged on a $\SI{4}{\milli\meter}$ grid, leading to a \SI{1.5}{\milli\meter} in between. At the bottom, details of the sensor and contact modeling are depicted as needed in \cref{sec:skin_model}. The discretization of the sensor is shown for an exemplary resolution of $\SI{0.5}{\milli\meter}$ per tactile point. Red tactile points correspond to the taxel regions $\mathcal{T}_j$, whereas blue points lie within the gaps and purple points form a margin around the sensor. These additional points are needed to correctly model the contact at each point of the sensor.}
    \label{fig:tactile_sensor}
\end{figure}

In this work we use a tactile skin glued to the soft fingertip of a multi-fingered hand (see \cref{fig:title_figure} and \cref{fig:tactile_sensor}). Although camera-based sensors like GelSight~\cite{Yuan.2017} and DIGIT~\cite{Lambeta.2020} provide a higher spatial resolution and are more widely used, they are bulky and can not be easily added to an existing robot structure (e.g., the complete fingertip hast to replaced). 

Important for all sensors is that they are soft so that in contact, the object can penetrate to some extent, resulting in a contact area and not only an uninformative point contact (\cref{fig:tactile_sensor}).
Regardless of the used sensor type, the challenge in simulating the spatially resolved tactile response is the need to model this soft contact. Vanilla rigid body simulators, which are the standard in reinforcement learning because of their high computational efficiency, do not suffice: in these simulators contacts between two colliding objects result only in a single (or few) contact point(s), i.e., no spatially resolved tactile image is provided.

\subsection{Related Work}
In this section, we discuss the existing work on tactile simulation and robotic fine manipulation.

\subsubsection*{Tactile Simulation}
The most accurate way for simulating the tactile sensor response would be to incorporate softness and deformations directly in the physics simulator. 
\citet{Narang.2021b,Narang.2021} used Finite element methods (FEM) for the (soft) BioTac sensor.
However, FEM-based simulations are infeasibly slow to be used in deep reinforcement learning.
\citet{Moisio.2013} simulate softness by evaluating a penalty-based contact model at a fixed set of points on a taxel-based sensor. 
The computed reaction forces at each point are used to compute the response of the tactile sensor but also to simulate the physics.
\Citet{JieXu.2023} use  a similar model to simulate the contact dynamics and sensor response of a vision-based sensor. 
Another custom soft-contact model for an optical tactile sensor is presented in \citet{Ding.2020}. 

Many recently developed tactile simulators (mainly for vision-based sensors) follow a more efficient approach by decoupling the physics simulation from the actual simulation of the tactile response. Typically, a fast rigid body simulation engine is used which feeds its output (object poses, penetration depth, forces, ...) to a more or less approximated soft contact model for computing the tactile response.
\citet{Gomes.2021} assume a known penetration of a manipulated object into the sensor, i.e., there is no dependence on the contact force. 
Other works assume compliant contacts~\cite{AlexChurch.2022,Lin.2022,Lin.2023}, but rely directly on the penetration depth from the rigid body simulation to compute the tactile response. 
While compliant contact forces linearly relate the penetration to the applied force, the penetration also depends on the number of contact points detected by the rigid-body simulator. This can lead to sudden jumps and makes the model unreliable.
The TACTO simulator~\cite{Wang.2022} avoids this problem, as the mapping between the force and penetration is separated from the physics simulation. 
This allows the use of non-linear mappings. 
However, this makes the penetration only dependent on the force but not on the contact geometry.
The penetration is observed by a camera within the simulated sensor and often an illumination model an/or a learned domain-adaptation is used to generate realistic images.

For taxel-based sensors, commonly multiple rigid bodies are used to simulate the tactile forces. 
One approach is to introduce softness by using compliant contacts for each taxel~\cite{Ding.2021,Yang.2023}. 
However, the simulation is not accurate enough such that the response of each taxel is reduced to a binary signal loosing sub-taxel precision. 
Alternatively, the softness can be introduced by using a mass-spring-damper systems~\cite{Habib.2014}. 
Using multiple bodies has the drawback that the contact geometry is altered leading to unnatural dynamics.

\subsubsection*{Fine Manipulation}
There exists only little work on learning fine manipulation of small objects in simulation with a successful sim-to-real transfer.
In \citet{Lambeta.2020}, a multi-fingered hand rolls a marble between two vision-based DIGIT sensors to a target position on the fingertip.
This task is closely related to our's (see \cref{fig:title_figure}) but the surfaces of the DIGITS are flat in contrast to the round fingertips in our setting, making it significantly easier to keep a stable contact. 
Moreover, the marble is rolled to single target positions while we aim to follow a continuous trajectory. 
Furthermore, they do not learn the rolling in simulation but train a dynamic model with real-world data and apply model predictive control. This approach was first used in \Citet{Tian.2019} but for a simpler setting: rolling a marble between a GelSight-style tactile sensor and a fixed flat surface making the task significantly easier.
\Citet{AlexChurch.2022} learn this easier task with simulation-based reinforcement learning and demonstrate a successful sim-to-real transfer.
\citet{Do.2023} focus especially on the inter-finger manipulation of small objects\ like screws, which are picked up from a bowl between two vision-based tactile sensors and are reoriented for a better classification. 
However, the reorientation is not learned but follows a handcrafted strategy.

\subsection{Contributions}
Our main contributions are as follows:

\begin{itemize}
\item A novel model for a tactile skin with soft contacts coupled to a regular rigid-body simulator that does not use the incorrect penetration depths from the simulator but computes for each tactile image the penetration depth from a force equilibrium.
\item A self-contained calibration procedure to tune the parameters of the tactile skin model. No external sensors or a test bench setup is needed, but only a cap with an indenter has to be mounted on a fingertip.
\item Two fine manipulation tasks, which highlight the need for tactile feedback, are presented: rolling a marble between two fingers on a specified trajectory and rolling an arbitrary oriented bolt along its normal direction. The tasks are solved via reinforcement learning using straightforward rewards. 
\item A detailed analysis with simulation experiments demonstrating the advantages of tactile feedback for fine manipulation and also that our skin model captures important effects allowing, e.g., sub-taxel precision.
\item Experiments on the real DLR-Hand~II demonstrate the successful zero-shot sim2real transfer. Although both policies directly use the raw tactile signal, the sim2real transfer works, proving the validity of our tactile skin simulation model.
\item To our best knowledge, the two shown fine manipulation tasks are the most challenging ever reported on a real robotic system.
\end{itemize}

\section{Simulating a Tactile Skin}
First, we present our general model idea. Afterward, we detail the implementation specific for the used TekScan Sensor and the fingertips of the DLR-Hand~II.
\subsection{Skin Model}
\label{sec:skin_model}
\subsubsection*{General Model Idea}
First of all, the surface of the sensor is discretized into a set of $N_t$ tactile points 
\begin{equation}
    \label{eq:tactile_point_set}
    \mathcal{T} = \left\{t_i = (\Vec{\boldsymbol{p}}_{t,i},\, \Vec{\boldsymbol{n}}_{t,i},\, a_{t,i}) \::\: i = 1, ..., N_t\right\},
\end{equation}
where $\Vec{\boldsymbol{p}}_{t,i} \in \mathbb{R}^3$ describes the Cartesian position, $\Vec{\boldsymbol{n}}_{t,i} \in \mathbb{R}^3$ the surface normal vector, and $a_{t,i} \in \mathbb{R}$ the surface area depending on the discretization grid (see \cref{fig:skin_model}, top left).
We assume that the indenter penetrates the fingertip by a maximum distance $\epsilon_\text{max}$ in the direction of the contact normal $\Vec{\boldsymbol{n}}_c$.
As a result, we can define a local penetration $\epsilon_i$ for every tactile point $t_i$ (see \cref{fig:skin_model}, top right).
We now assume a purely linear relationship between the local penetration and sensor response $\sigma_i = E \epsilon_i$, where the factor E is a tunable parameter of the model.
The notation was chosen to resemble Hooke's law: $\sigma_i$ describes the stress at the fingertip in the direction of the contact normal, and $E$ is comparable to a modulus of elasticity.

Each taxel of the sensor patch consists of a subset of all tactile points: 
\begin{equation}
    \mathcal{T}_j \subset \mathcal{T},\: j = 1, ..., N_T,
\end{equation}
where $N_T$ is the number of taxels of the sensor.
Therefore, the force applied to the region of each taxel can be calculated by summing the normal stresses $\sigma_i$ of all the respective tactile points:
\begin{equation}
    F_j = \sum_{t_i \in \mathcal{T}_j} \sigma_i a_{t,i} \max\left(\Vec{\boldsymbol{n}}_c \cdot \Vec{\boldsymbol{n}}_{t,i},\, 0\right),\; j = 1, ..., N_T. 
\end{equation}
The additional factor $\max\left(\Vec{\boldsymbol{n}}_c \cdot \Vec{\boldsymbol{n}}_{t,i},\, 0\right)$ projects the area of every tactile point onto the normal direction.
The lower bound of zero excludes tactile points from the sum that face in the opposite direction of the contact normal.
Moreover, we assume that the sensor only senses forces in its local normal direction but no shear forces.
If the indenter is rigid, the elastic fingertip will bend around it.
The stresses $\sigma_i$ act in the direction of the contact normal but the normal of the fingertip surface is orientated as the surface normal of the indenter (see \cref{fig:skin_model}, bottom left). 
Therefore, an additional projection factor is introduced. 
For every tactile point, the stress is projected on the respective surface normal: $(-\Vec{\boldsymbol{n}}_{s,i} \cdot \Vec{\boldsymbol{n}}_c)$.
Finally, the taxel values $T_j$ of the real sensor lie between $0$ and $255$.
Hence, an additional, taxel-specific scaling parameter $S_j$ is introduced.
Eventually, the sensor model to compute the value of every taxel is given as
\begin{equation}
    T_j = S_j \sum_{t_i \in \mathcal{T}_j} E \epsilon_i a_{t,i} (\Vec{\boldsymbol{n}}_c \cdot \Vec{\boldsymbol{n}}_{t,i}) (-\Vec{\boldsymbol{n}}_{s,i} \cdot \Vec{\boldsymbol{n}}_c),\; j = 1, ..., N_T.
\end{equation}

\subsubsection*{Interplay with Rigid-Body Simulator}
As mentioned above, the model is designed to work on top of a regular rigid-body simulator.
Hence, the maximum penetration depth is not modeled correctly but must be computed separately. 
The simulator provides the relative position of the objects, the contact normal $\Vec{\boldsymbol{n}}_c$, and the normal force $F_n$. The local penetrations $\epsilon_i$ depend solely on the maximum penetration $\epsilon_\text{max}$, which can be varied by moving the relative position along the direction of the provided contact normal. The maximum penetration is chosen such that the sum of all local stresses matches the reported normal force:
\begin{equation}
    \label{eq:force_equilibrium}
    F_n \overset{!}{=} \sum_{t_i \in \mathcal{T}} E\epsilon_i(\epsilon_\text{max}) a_{t,i} (\Vec{\boldsymbol{n}}_{t,i} \cdot \Vec{\boldsymbol{n}}_c).
\end{equation}

\subsubsection*{Obtaining the Local Penetrations}
\begin{figure}
    \centering
    \includegraphics[width=1.0\linewidth]{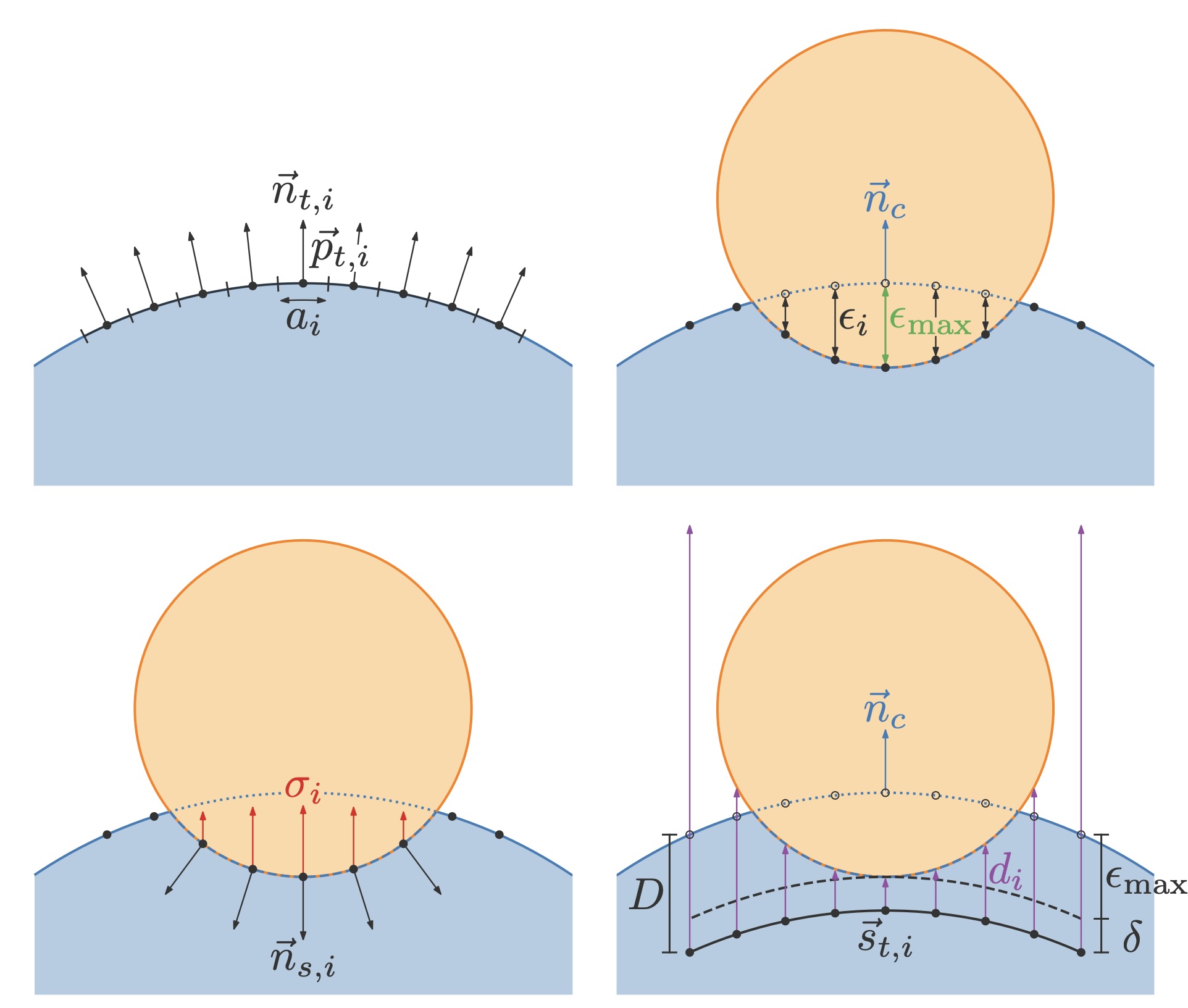}
    \vspace{-0.6cm}
    \caption{Visualization of the tactile skin model as detailed in \cref{sec:skin_model}.}
    \vspace{-0.5cm}
    \label{fig:skin_model}
\end{figure}

The local penetration depths $\epsilon_i$ are computed using an operation called ray casting, which determines the distance in a certain direction between a point in space and a mesh object (see \cref{fig:skin_model}, bottom right).
First of all, the contact between the fingertip and the indenter is resolved by moving the tactile points along the contact normal by a large enough constant $D$ so that they all lie outside of the indenter:
\begin{equation}
    \Vec{\boldsymbol{s}}_i = \Vec{\boldsymbol{p}}_{t,i} - D\Vec{\boldsymbol{n}}_c,\; i = 1, ..., N_t.
\end{equation}
From these shifted points, a ray cast in the direction of the contact normal determines the distances $d_i$ to the indenter.
The offset $\delta$ between the starting points and the deepest point of the mesh can be computed as $\delta =\min d_i$. The local penetrations are then given by subtracting the measured distances $d_i$, corrected by the offset $\delta$:
\begin{equation}
    \epsilon_i(\epsilon_\text{max}) = \max\left(\epsilon_\text{max} - (d_i - \delta),\, 0\right),\; i=1,...,N_t.
\end{equation}
The maximum operation is required to account for tactile points that are not penetrated by the indenter, which would otherwise lead to negative penetration depths. 
The ray casting itself is independent of the maximum penetration as it only occurs in the formula for the local penetrations.
Hence, the expensive ray cast operation must only be executed once before searching the maximum penetration satisfying the equilibrium defined in \cref{eq:force_equilibrium}.

\subsection{Model Implementation}
The tactile skin is bent around the cylindrical part of the fingertip. 
The position can be described by three parameters:
\begin{itemize}
    \item $y$: The offset along the cylinder's longitudinal axis, measured as a distance.
    \item $\beta$: The offset along the perimeter of the cylinder, measured as an angle around the longitudinal axis.
    \item $\alpha$: The orientation of the sensor on the cylinder, measured as an angle around the surface normal of the cylinder.
\end{itemize}
The discretization of the sensor is shown in the bottom right of \cref{fig:tactile_sensor}.
For our implementation we use a resolution of $\SI{0.25}{\milli\meter}$ per taxel and a margin of $\SI{8}{\milli\meter}$ leading to $\SI{14884}{}$ tactile points. 
We incorporate our model into the rigid-body simulator PyBullet \cite{Coumans.2016}, where we simulate two fingers of the DLR-Hand~II. 
For computing the ray casts, we rely on the Python library Trimesh \cite{trimesh}. 
Finding the correct $\epsilon_\text{max}$ is done via a line search with a step size of $\SI{0.1}{\milli\meter}$. 
While the physics simulation runs at a rate of $\SI{1000}{\hertz}$, the soft contact model for the tactile images is evaluated every $100th$ step. 
The simulation of $\SI{0.1}{\second}$ takes in total $\SI{56}{\milli\second}$, where the computation of the tactile images takes $\SI{26}{\milli\second}$;
i.e., the simulation runs with a real-time factor of $1.8$ on a small single-core AMD EPYC 7B13 @ $\SI{3}{\giga\hertz}$. 
The bottleneck is the expensive ray cast operation as the line search takes only $\SI{0.3}{\milli\second}$.

\section{Model Calibration}
To transfer control policies learned in simulation to the real-world system, the simulation must resemble reality as accurately as possible. 
Therefore, this chapter presents a calibration procedure for the model parameters. 
A perfect calibration is not required for a successful sim-to-real transfer, but identifying reasonable parameter intervals for domain randomization is sufficient.
\subsection{Self-Contained Data Generation}
\begin{figure}[!htb]
    \centering
    \includegraphics[width=0.95\linewidth]{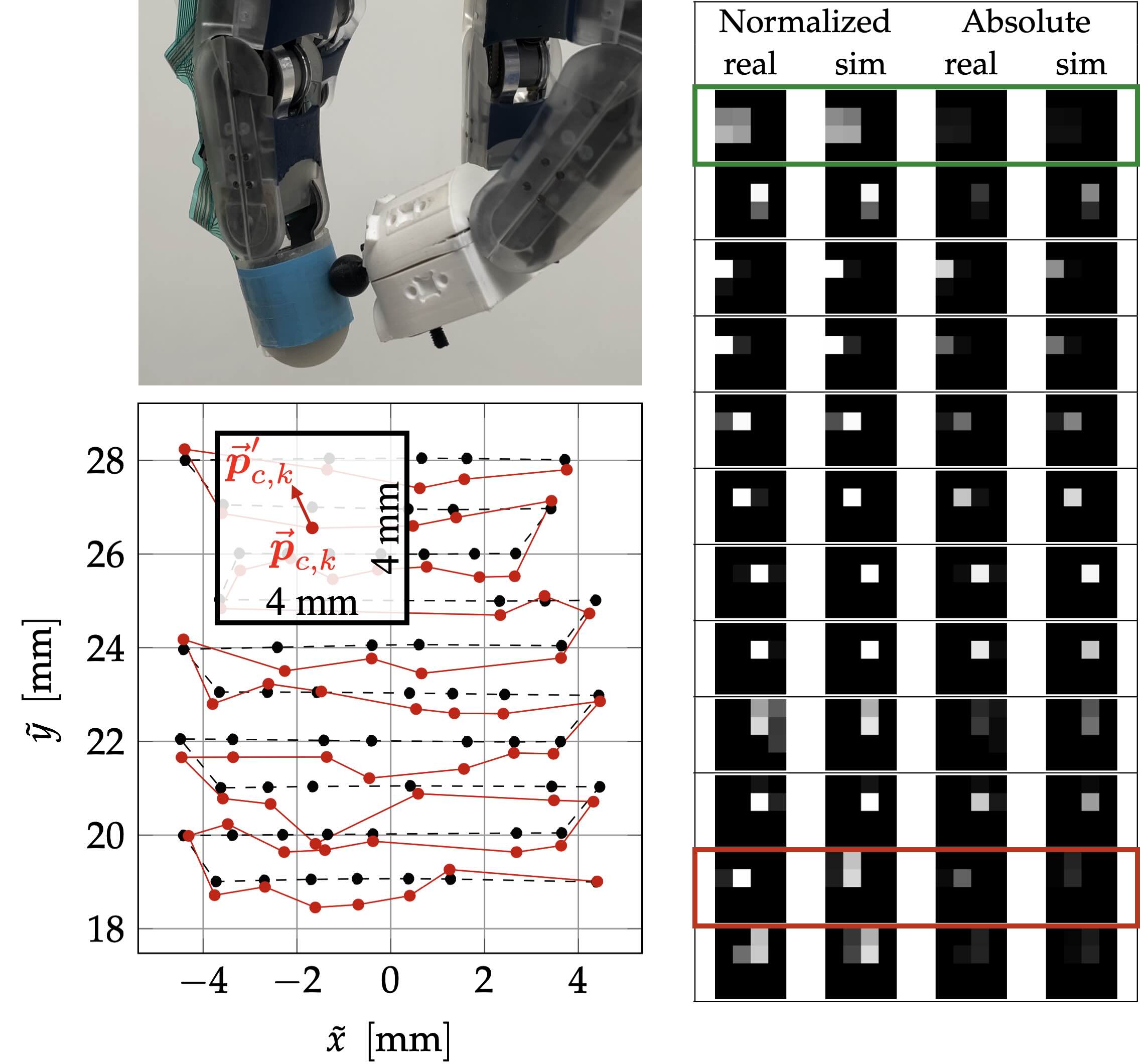}
    \vspace{-0.3cm}
    \caption{\textbf{Left:} Self-Contained data generation. The top image shows the cap with an spherical indenter of radius $\SI{6}{\milli\meter}$. The bottom image shows the set of $60$ desired contact points, randomly selected from a uniform grid with a resolution of $\SI{1}{\milli\meter}$. While the desired contact points are shown in black, the measured contact points are shown in red. Moreover, the slack introduced for each measured contact point is visualized.
    \\\textbf{Right:} Results of the micro variable optimization for an exemplary set of macro variables. The green box marks a positive result where the images match well between simulation and reality. The red box marks a negative example that can be caused if the slack of $\pm\SI{2}{\milli\metre}$ is not sufficient.}
    \label{fig:indenter_cap}
\end{figure}

A cap with a spherical indenter is mounted to the thumb and the indenter is pressed against the tactile skin attached to the forefinger at a set of desired contact points (see \cref{fig:indenter_cap}). 
At each point, the desired contact force is sampled uniformly between $\SI{1}{\newton}$ and $\SI{4}{\newton}$ and is applied by commanding the required torques to the fingers.
The measured contact points $\Vec{\boldsymbol{p}}_{c,k}$ can be calculated with the forward kinematics of the fingers and the measured joint angles. 
Moreover, the real contact force $F_{n,k}$ can be reconstructed by the measured joint torques. 
We want to calibrate the skin at sub-taxel precision. 
Hence, we keep only samples where at least two taxels were active in the measured tactile image $\boldsymbol{T}_k$ during contact. 
Eventually, the data set includes $N_P=24$ valid samples including the measured contact positions, contact forces, and tactile images: 
\begin{equation}
    \mathcal{D} = \left\{(\Vec{\boldsymbol{p}}_{c,k},\, F_{n,k},\, \boldsymbol{T}_k)\::\: k = 1, ..., N_P\right\}.
\end{equation}

\subsection{Model Fitting as an Optimization Problem}
The goal of the calibration is to optimize the model parameters so that the simulated tactile images best match the measured ones:
\begin{equation}
    \argmin_{E,\, S,\, y,\, \beta,\, \alpha}\;\underbrace{\frac{1}{16N_V}\sum_k\lVert\boldsymbol{T}(\Vec{\boldsymbol{p}}_{c,k},\, F_n,\, E,\, S,\, y,\, \beta,\, \alpha) - \boldsymbol{T}_k\rVert_2^2}_{\mathcal{L}_\text{mse}},
\end{equation}
where each simulated image is generated for the measured contact position and force.

However, small errors in the measured contact position due to an imperfect kinematic model of the fingers can cause significantly different tactile images. 
Although we calibrated the kinematic model \cite{Tenhumberg.2023}, the remaining uncertainty of a few millimeters is still too high as the taxels measure only $\SI{2.5}{\milli\meter}$.
Hence, we additionally introduce the estimated contact positions $\Vec{\boldsymbol{p}}_{c,k}'$ to the optimization problem that allow a slack of $\pm\SI{2}{\milli\meter}$ around each measured contact point $\Vec{\boldsymbol{p}}_{c,k}$ (see \cref{fig:indenter_cap}, bottom left).

\begin{figure}
    \centering
    \includegraphics[width=0.85\linewidth]{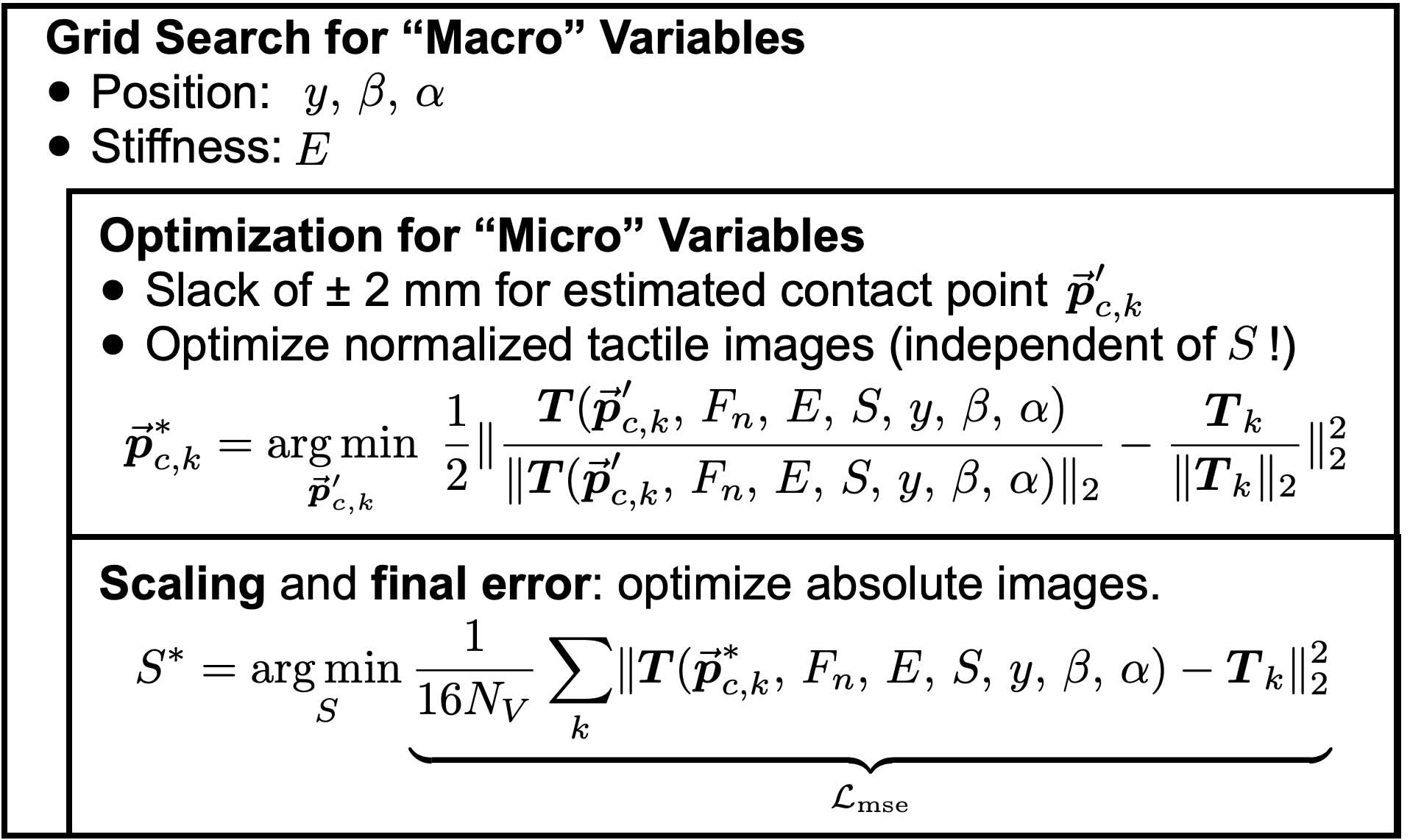}
    \vspace{-0.3cm}
    \caption{Calibration scheme.}
    \label{fig:calibration_scheme}
\end{figure}

To solve the optimization problem, it is divided into two stages as introdcued in \cref{fig:calibration_scheme}.
For a given set of macro variables, the estimated contact positions are optimized using the normalized tactile images, because the information on the contact position lies purely in the ratio of the taxel values but not in their absolute values.
The optimal scaling can then be obtained analytically by solving the least squares problem for the absolute images.
The results of an exemplary micro variable optimization are shown on the right side of \cref{fig:indenter_cap}.

Due to the slack introduced by the micro variables, the grid search results do not have a sharp minimum. 
Instead they show a valley with a loss of $\mathcal{L}_\text{mse}\leq25$, i.e., a root mean square error of less than 5 per taxel.
We take the boundaries of this valley as the intervals for the calibration result:
\begin{itemize}
    \item $y \in [\SI{21.75}{\milli\meter};\; \SI{25.25}{\milli\meter}]$,
    \item $\beta \in [\SI{0}{^\circ};\; \SI{5.03}{^\circ}]$,
    \item $\alpha \in [\SI{-15}{^\circ};\; \SI{10}{^\circ}]$,
    \item $E \in [\SI{236}{\mega\pascal\per\meter};\; \SI{848}{\mega\pascal\per\meter}]$.
\end{itemize}
The scaling factors corresponding to all macro variable combinations with $\mathcal{L}_\text{mse}\leq25$ form a coherent cluster. Again, we take the boundary values to obtain the interval of reasonable values:
\begin{itemize}
    \item $S \in [\SI{61}{\per\newton};\; \SI{76}{\per\newton}]$.
\end{itemize}

\section{Application to Learning Fine Manipulation}
We want to learn two challenging fine-manipulation tasks in simulation. The first is rolling a marble along arbitrary trajectories between the fingers. 
The second is rolling a bolt following a sinus motion in different orientations.
The policy has to detect the orientation and roll the bolt in the according direction. 
To learn these tasks, we first introduce the simulation setup before specifying the learning details.

\subsection{Simulation and Environment Setup}
\begin{figure}
    \centering
    \includegraphics[width=\linewidth]{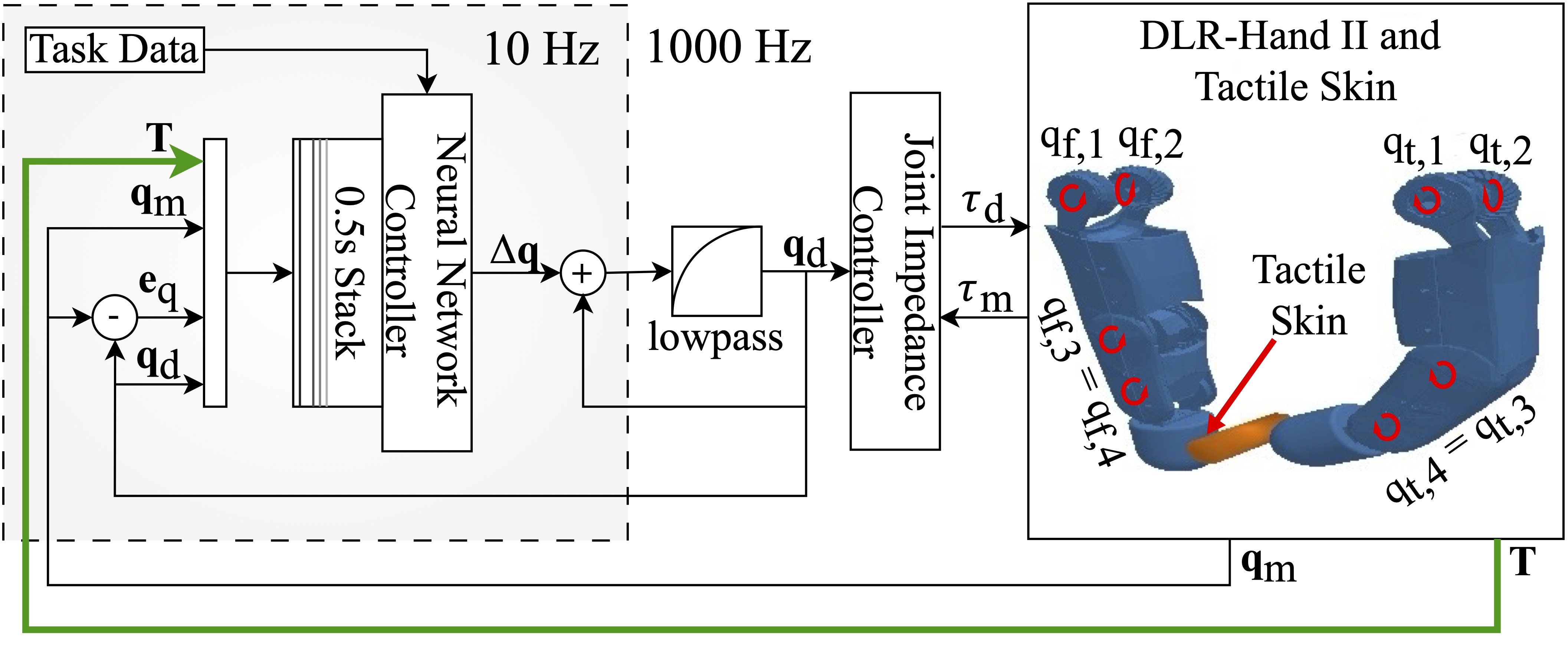}
    \vspace{-0.6cm}
    \caption[Control architecture.]{Embedding of the policy into the overall control architecture. The input of the neural network is a subset of the stack of the last five observations ($\boldsymbol{T}\in\mathbb{R}^{16}$ and $\boldsymbol{q}_m,\boldsymbol{q}_d,\boldsymbol{e}_q\in\mathbb{R}^6$), and task specific data. The output is a six-dimensional change of the desired joint angles.}
    \label{fig:control_architecture}
\end{figure}

The relevant fingers of the DLR-Hand~II and the manipulated object are loaded into the PyBullet simulator. 
The model and simulation of the torque-controlled fingers follows the work of \citet{Sievers.2022}. 
The fingers start already in a pinch grasp configuration and the object is placed between the finger tips. 
The focus of the learning is to utilize the tactile signal and not finding the correct grasp. 
For the bolt rolling task, the initial orientation $\alpha_b$ is randomly selected from the set $\{\SI{-45}{^\circ},\,\SI{0}{^\circ},\,\SI{45}{^\circ},\,\SI{90}{^\circ}\}$ at the beginning of each episode.
The manipulated object is initially kept in place with a position constraint. 
This constraint is removed when the fingers grasped the object successfully, i.e., both fingers are in contact. 
When the object is dropped, i.e., one of fingers lost contact, the current episode is terminated in failure, the simulation is reset to its initial state, and a new episode begins. 
Furthermore, we limit the duration of each episode to $\SI{6}{\second}$. 
After this timeout, the environment is reset without sending a termination signal to the reinforcement learning algorithm.
The control rate of the real robotic system is $f_\text{sys} = \SI{1000}{\hertz}$. 
Hence, we chose the same rate for the physics simulation. 
In contrast, the control policy is evaluated every $100^\text{th}$ environment step leading to a controller sample rate of $f_\text{cont} = \SI{10}{\hertz}$. 
The overall control architecture is summarized in \cref{fig:control_architecture}.
As the soft contact model for the tactile images, does not influence the physics simulation but only provides the tactile images, it is evaluated at the same rate as the policy.

\subsubsection*{Action Space}
The final output of the controller must be a vector of desired joint angles $\boldsymbol{q}_d \in \mathbb{R}^6$ that are commanded to the underlying joint impedance controller of the thumb and forefinger. The output of the policy $\boldsymbol{a}\in[-1;\; 1]$ is added to the latest desired joint angles. This corresponds to commanding a joint velocity and, hence, the output $\boldsymbol{a}$ is scaled to not exceed the maximum joint velocity $\Dot{q}_\text{max}$. Moreover, the output is clipped to stay within the joint limits:
\begin{equation}
    \boldsymbol{q}_\text{out} = \clip\left(\boldsymbol{q}_d + \Delta\boldsymbol{q},\, \boldsymbol{q}_\text{min},\, \boldsymbol{q}_\text{max}\right),\quad \Delta\boldsymbol{q} = \boldsymbol{a}\frac{\Dot{q}_\text{max}}{f_\text{cont}}. 
\end{equation}
Finally, to avoid jumps in the commanded signal, the commanded joint angles are smoothed by a first-order low-pass filter running at $f_\text{sys}$.

\subsubsection*{Observation Space}
The joints of the DLR-Hand~II are torque-controlled. Hence, a difference between the desired $\boldsymbol{q}_d$ and measured joint angles $\boldsymbol{q}_m$ exists that includes the information on the current torque: $\boldsymbol{\tau}=K(\boldsymbol{q}_d-\boldsymbol{q}_m)$, where $K$ is the stiffness of the impedance controller. 
Therefore, the control error $\boldsymbol{e}=(\boldsymbol{q}_d-\boldsymbol{q}_m)$ includes already some tactile information.  
Most robotic hands, however, do not have torque sensors. 
Hence, their controller can only set the desired joint angles. 
Any elasticity results purely from the mechanical structure but cannot be measured. 
This can be imitated by passing only the desired joint angles to the policy.
We train two policies for each task with (\textit{Tactile}) and without (\textit{Desired}) the skin feedback $\boldsymbol{T}$ in addition to the desired joint angles. 
To demonstrate the effectiveness of the tactile skin, we compare it to a third policy (\textit{Joint Angles}) that can leverage the torque information, which is specifically available for the DLR-Hand~II. 
The different observation spaces are summarized in \cref{tab:observation_spaces}. 
To further enrich the information, the aforementioned observations are stacked over the last five iterations so that the policy sees a history of $\SI{0.5}{\second}$.
Moreover, all policy receive the desired position of the manipulated object on the sensor $\Vec{\boldsymbol{p}}_{d,k}$ and the difference to the previous iteration $\Delta\Vec{\boldsymbol{p}}_{d,k}=\Vec{\boldsymbol{p}}_{d,k}-\Vec{\boldsymbol{p}}_{d,k-1}$. 
For the marble, this is a two-dimensional vector as it can be an arbitrary position on the fingertip. 
In contrast, for the bolt it is one-dimensional as it only specifies the position of the up-down movement but the orientation must be figured out by the policy itself.

\begin{table}
    \caption{Observation space of different policies.}
    \label{tab:observation_spaces}
    \centering
    \begin{tabular}{lcccc}
        \hline
        \multicolumn{2}{c|}{Observation} & Desired & Joint Angles & Tactile \\\hline
        Tactile image & $\boldsymbol{T}$ &  &  & $\checkmark$  \\
        Desired joint angles & $\boldsymbol{q}_d$ & $\checkmark$  & $\checkmark$  & $\checkmark$  \\
        Measured joint angles & $\boldsymbol{q}_m$ & & $\checkmark$  & \\
        Control error & $\boldsymbol{e}_q$ &  & $\checkmark$  &\\\hline
        Target position & $\Vec{\boldsymbol{p}}_{d,k}$ & $\checkmark$  & $\checkmark$  & $\checkmark$ \\
        Target movement & $\Delta\Vec{\boldsymbol{p}}_{d,k}$ & $\checkmark$  & $\checkmark$  & $\checkmark$ 
    \end{tabular}
\end{table}

\subsection{Task Definition}
To formalize the two tasks, we have to define a reward function for the reinforcement learning algorithm. The individual components are presented below.

\subsubsection*{Safety Critical Rewards}
To prevent self-destructive behavior, we introduce penalty terms for getting close to the physical limitations of the hand. 
For the penalty terms we use the following function:
\begin{equation}
    \pen(x;\, \Delta x) = \left\{
    \begin{array}{ll} 
        0, & x < 0 \\
        \frac{1}{2}\left(1-\cos{\left(\frac{x}{\Delta{x}}\pi\right)}\right), & 0 \leq x \leq \Delta{x} \\
        1, & x > \Delta{x}.
    \end{array}\right.
\end{equation}
We penalize running into joint, joint velocity, and torque limits, where $x_{q/\Dot{q}/\tau}$ describe how much the values exceed the soft margins, which are $\Delta x_{q/\Dot{q}/\tau}$ away from the hard constraints:
\begin{equation}
    p_{q/\Dot{q}/\tau} = \lambda_{q/\Dot{q}/\tau}\pen(x_{q/\Dot{q}/\tau};\, \Delta x_{q/\Dot{q}/\tau}).
\end{equation}

\subsubsection*{Holding Rewards}
The focus of the learning is to utilize the skin signal. Hence we define two rewards pointing the policy towards the correct grasp. 
The first just gives a binary reward for touching the object with the tip of the thumb and the cylindrical part of the forefinger:
\begin{equation}
    r_g = \left\{
    \begin{array}{ll} 
        0, & \text{no or incorrect contact} \\
        \lambda_g, & \text{correct contact.}
    \end{array}\right.
\end{equation}
The second reward is granted for holding the object with a normal force $F_n$ of roughly $\SI{1.5}{\newton}$: $r_f=\lambda_f\rew(|F_n-\SI{1.5}{\newton}|;\, \SI{1.5}{\newton})$, where the general shape of the reward function is given as
\begin{equation}
    \rew(x;\, \Delta x) = \left\{
        \begin{array}{ll} 
            0, & x < 0 \\
            \frac{1}{2}\left(1+\cos{\left(\frac{x}{\Delta{x}}\pi\right)}\right), & 0 \leq x \leq \Delta{x} \\
            1, & x > \Delta{x}.
        \end{array}\right.
\end{equation} 

\subsubsection*{Marble Rolling}
To learn rolling the marble along arbitrary trajectories, we reward the policy for keeping the measured contact point $\Vec{\boldsymbol{p}}_k$ of the marble on the fingertip close to the desired position $\Vec{\boldsymbol{p}}_{d,k}$:
\begin{equation}
    r_p = \lambda_p\rew(\lVert\Vec{\boldsymbol{p}}_{d,k}-\Vec{\boldsymbol{p}}_{k}\rVert_2, \SI{5}{\milli\meter}).
\end{equation}
The trajectories of the desired positions are generated randomly during training and move within a $\SI{6}{\milli\meter}\times\SI{6}{\milli\meter}$ box.
The positions are defined relative to the sensor position. 
Eventually, the final reward for rolling the marble is given as
\begin{equation}
    r = r_p + r_f + r_g - p_q - p_{\Dot{q}} - p_\tau.
\end{equation}

\subsubsection*{Bolt Rolling}
For the bolt rolling task, we reuse the location reward from the marble rolling. 
However, the desired position provided to the policy describes only the position along one-dimensional sinus movement. 
Hence, the two-dimensional position vectors required by the reward are computed using the initial orientation of the bolt $\alpha_b$. 
The motions have an amplitude of $\SI{6}{\milli\meter}$ and the frequency is sampled between $\SI{0.25}{\hertz}$ and $\SI{0.5}{\hertz}$.
Again the positions are specified relative to the sensor position.
Moreover, an additional reward for keeping the bolt in its initial orientation relative to the sensor is introduced:
\begin{equation}
    r_\alpha = \lambda_\alpha\rew(|\alpha_{b}-\alpha_{b,k}|, \SI{15}{^\circ}).
\end{equation}
Hence, the final reward for the bolt rolling task is given as
\begin{equation}
    r = r_p + r_\alpha + r_f + r_g - p_q - p_{\Dot{q}} - p_\tau.
\end{equation}

\subsection{Domain Randomization}
For a successful Sim2Real transfer we apply domain randomization during simulation. 
See \cref{tab:domain_randomization} for the most important parameters and the website for a complete list.

\begingroup
\setlength{\tabcolsep}{3pt}
\begin{table}
    \caption{Domain Randomization.}
    \label{tab:domain_randomization}
    \centering
    \begin{tabular}{lcccc}
        \hline
        \multicolumn{2}{c}{Parameter} & Unit & Distribution & Mode  \\\hline
        Joitn offsets\footnotemark & ${q}_\text{off}$ & $\si{\radian}$ & $\mathcal{U}(-0.04,\, 0.04)$ & per episode \\
        Joitn noise & ${q}_\text{noise}$ & $\si{\radian}$ & $\mathcal{N}(0,\, 0.02)$ & per step \\\hline
        \multirow{3}{*}{Sensor position} & $y$ & $\si{\milli\meter}$ & $\mathcal{U}(21.5,\, 25.5)$ & per episode \\
        & $\beta$ & $\si{^\circ}$ & $\mathcal{U}(-12,\, 12)$ & per episode \\
        & $\alpha$ & $\si{^\circ}$ & $\mathcal{U}(-15,\, 15)$ & per episode \\
        Elasticity & $E$ & $\si{\mega\pascal\per\meter}$ & $\mathcal{U}(236,\, 848)$ & per episode\\
        Scale & $S_j$ & $\si{\per\newton}$ & $\mathcal{U}(61,\, 76)$ & per episode \\
        Taxel offsets & $T_\text{off}$ & $1$ & $\mathcal{U}(-5,\, 5)$ & per episode  \\
        Noise range & $\sigma_\text{taxel}$ & $1$ & $\mathcal{U}(0,\,5)$ & per episode \\
        Taxel noise & $T_\text{noise}$ & $1$ & $\mathcal{N}(T_\text{off},\, \sigma_\text{taxel})$ & per step 
    \end{tabular}
\end{table}
\footnotetext{Just before publication, we noticed that we sampled the joint offsets not independently -- but this amplifies the domain randomization even more.}
\endgroup

\subsubsection*{Hand Parameters}
To introduce the uncertainties of the kinematic model we sample joint offsets $q_\text{off}$ at the beginning of each episode that are added to the measured joint angles to imitate a constant error. 
Moreover, Gaussian noise is added to the measured joint angles in each step to simulate, for instance, slack in the joint gears.
Furthermore, domain randomization is applied to the controller parameters.

\subsubsection*{Object parameters}
While the bolt keeps a constant radius of $\SI{6}{\milli\meter}$, we vary the radius of the marble from $\SI{4}{\milli\meter}$ to $\SI{8}{\milli\meter}$. 
Domain randomization is also applied to the dynamic parameters such as mass or friction. 
Moreover, noise is added to the initial position and orientation of the object at the beginning of each episode.

\subsubsection*{Skin parameters}
The intervals for the domain randomization of the scaling and elasticity are taken from the calibration results. 
The sensor position is, however, based on the precision that can be achieved by manually gluing the sensor to the fingertip, as it is repeatedly removed between experiments. 
To allow a fair comparison, the sensor position is not randomized for policies that do not use the tactile feedback but fixed at the middle of the fingertip. 
The target trajectories are defined relative to the sensor position and without tactile feedback the policies cannot know the skin position. 
Hence, it is kept constant.
Moreover, noise is added to the tactile images to better resemble the real sensor.
At the beginning of each episode, constant offsets and noise ranges are sampled for each taxel. 
During the episodes, Gaussian noise with a mean of the offset and a standard deviation of the noise range is added to the tactile images.

\subsection{Reinforcement Learning Training}
To implement the simulation environment, we stick to the interface defined by the Gymnasium API standard \cite{Brockman.2016}. 
This makes the environment compliant with most reinforcement learning frameworks. 
We use the Soft Actor Critic \cite{Haarnoja.2018} algorithm implemented in Stable Baselines3 \cite{Raffin.2021} to train the policies. 
We rely on simple multilayer perceptrons with two hidden layers for the actor and critic networks.
Each hidden layer has $256$ neurons, and ReLU activation functions are used. 
All network inputs are normalized, with the means and variances of all input dimensions obtained as a running average during training. 
The networks are trained with the Adam optimizer \cite{Kingma.22.12.2014}.
We use asymmetric observations \cite{Andrychowicz.2020, OpenAI.2019, Sievers.2022, Pitz.2023} to provide additional information to the critic networks. 
We train the marble policies for $\SI{800000}{}$ and the bolt policies for $\SI{600000}{}$ environment steps, while using $8$ parallel environments.

\section{Results}
We first highlight the importance of tactile feedback by analyzing the marble rolling policy in simulation. 
Afterward, we present the successful sim-to-real transfer of the policies.

\subsection{Importance of Tactile Feedback}
\subsubsection*{Marble}
\begin{figure}
    \centering
    \includegraphics[width=0.9\linewidth]{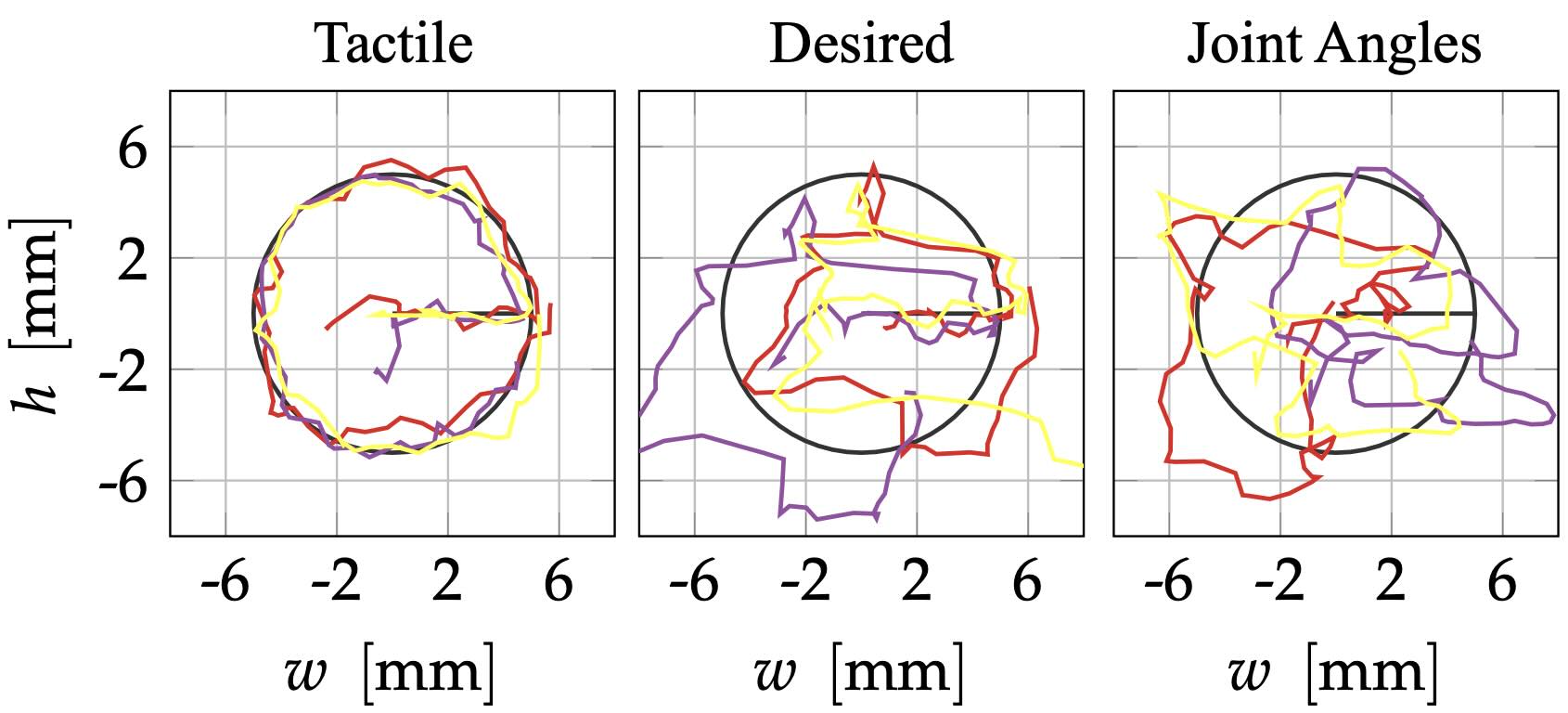}
    \vspace{-0.3cm}
    \caption{Qualitative analysis of marble rolling. Three exemplary runs are shown for each policy. While the circle is tracked smoothly with tactile feedback, the trajectories without it follow the target position poorly and the circular pattern can only be recognized roughly.}
    \label{fig:marble_examples}
\end{figure}

\begin{figure}
    \centering
    \includegraphics[width=0.7\linewidth]{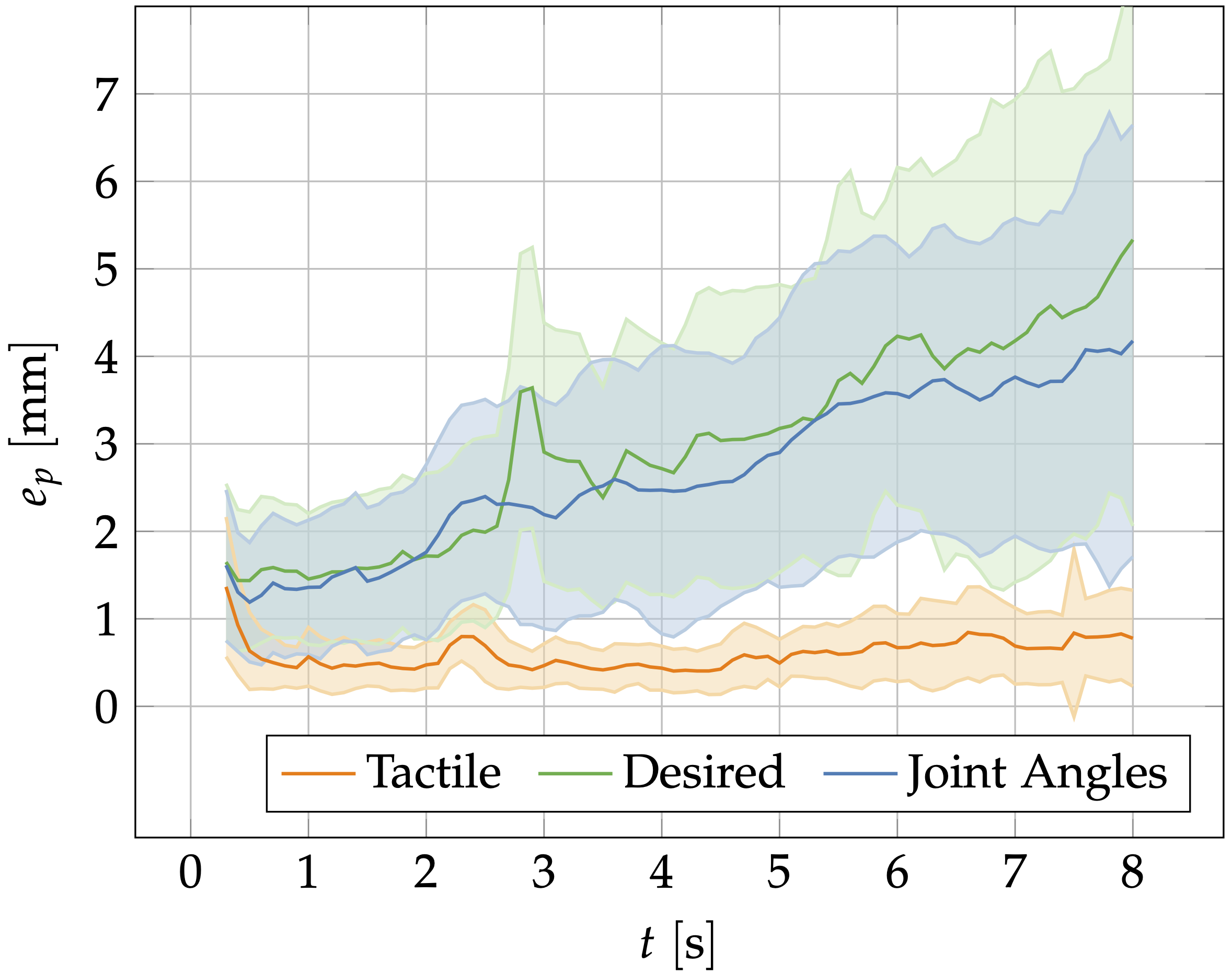}
    \vspace{-0.3cm}
    \caption{Quantitative analysis of marble rolling. The plot shows the tracking error $e_p$ over time for each policy. With tactile feedback it reduces from initially $\SI{1.5}{\milli\meter}$ to approximately half a millimeter. 
    After $\SI{4.5}{\second}$, the error marginally increases as the manipulation makes the pinch grasp more and more unstable over time. 
    After $\SI{8}{\second}$ the error is approximately $\SI{0.8}{\milli\meter}$. For the policies w/o tactile feedback, the tracking error monotonically increases from $\SI{1.5}{\milli\meter}$ to $\SI{5}{\milli\meter}$ (\textit{Desired}) and $\SI{4}{\milli\meter}$ (\textit{Joint Angles}) after $\SI{8}{\second}$.}
    \label{fig:marble_analysis}
    \vskip -0.5cm
\end{figure}

To evaluate the performance of the different marble rolling policies, a circular trajectory with a radius of $\SI{5}{\milli\meter}$ and a duration of $\SI{6}{\second}$ is commanded after moving from the center to the side for $\SI{2}{\second}$.
During the evaluation the domain randomization, such as the sensor position, is turned off to allow a fair comparison between the policies with and without tactile feedback.
Just the randomization of the joint offset is left active to keep the effect of the kinematic uncertainties.
The results are analyzed qualitatively in \cref{fig:marble_examples} and quantitatively in \cref{fig:marble_analysis}.
Recall that the taxels measure $\SI{2.5}{\milli\meter}$, i.e., the marble is manipulated at sub-taxel precision with tactile feedback as the tracking error is less than a millimeter.
The feedback of the measured joint angles and the control error that include the tactile information of the torque sensors only marginally improves the performance compared to the policy receiving only the desired joint angles. 
Hence, tactile feedback is essential for the precise fine manipulation of small objects.


\subsubsection*{Bolt}
To roll the bolt in the correct direction, the policy must estimate its orientation. 
During training we observe that this is only possible with tactile feedback but not with the desired joint angles alone. 
With access to the measured joint angles the policy can infer the orientation by "wiggling motions". 
However, this is limited to special robotic hands with precise torque control such as the DLR-Hand~II.

\subsection[title]{Simulation to Reality Transfer\footnote{To compensate for sensitivity loss due to wear and tear, before the shown experiments an overall scaling factor for the taxel response was determined by rolling the bolt on the real robot and in simulation.}}
For detailed results of the sim-to-real transfer, please refer to the accompanying video.
\subsubsection*{Marble}
\begin{figure}
\centering	
    \includegraphics[width=0.7\linewidth]{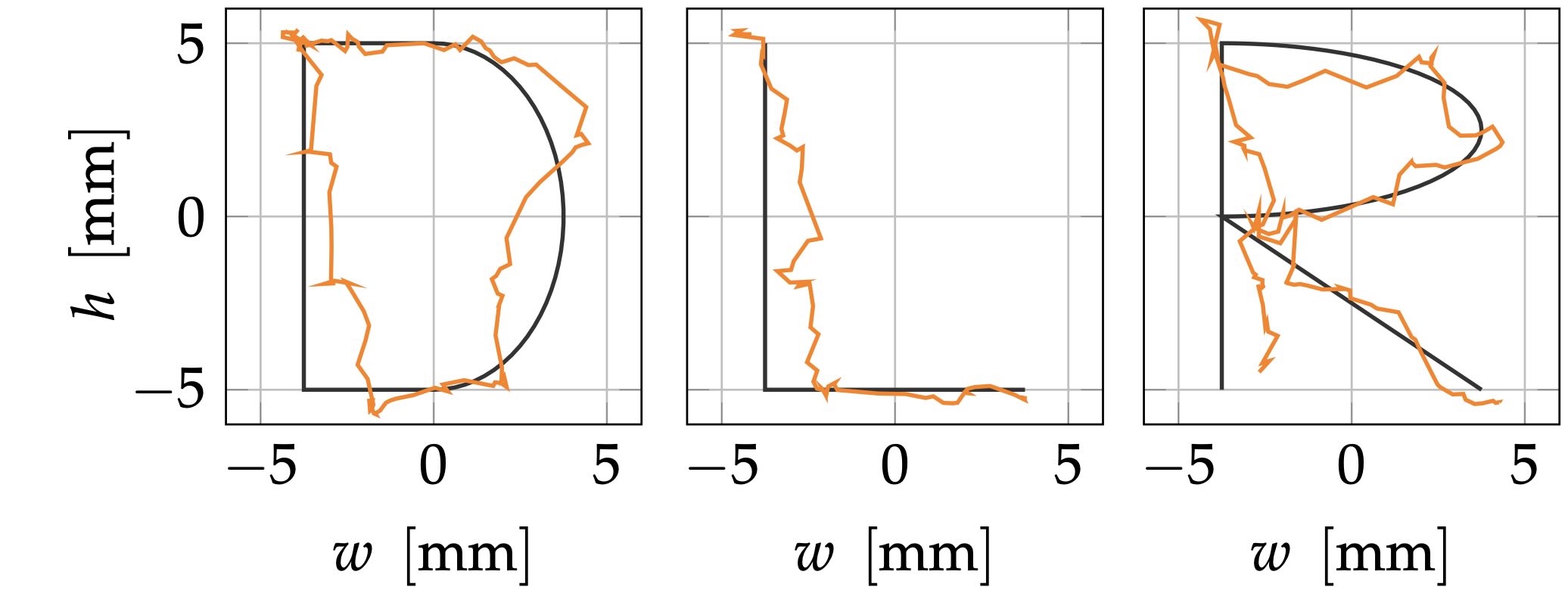}
\vspace{-0.3cm}
    \caption{Writing letters with the marble in reality. The desired trajectory of each letter is shown in black and the estimated position of the marble is shown in orange. The estimated position is obtained by training a neural network to predict the marble position based on the tactile images. As no real-world test data is available, the result is only qualitative. The run is shown in the accompanying video.}
    \vskip -0.2cm
    \label{fig:letters}
\end{figure}

To test the marble rolling policy we command the same trajectory as for the simulation analysis except we add two more circles so that one run takes $\SI{20}{\second}$.
Each policy was executed three times for a marble with a radius of $\SI{5}{\milli\meter}$ and $\SI{7}{\milli\meter}$, respectively. 
Although all three policies achieved a success rate of $\SI{100}{\percent}$, the tactile policy tracked the circle significantly smoother than the ones without tactile feedback. 
An exemplary run is shown in \cref{fig:title_figure}.
To push the tactile policy to its limits, we choose a trajectory that depicts the initials of the German Aerospace Centre (DLR). 
The letters are "written" on the sensor by the marble sequentially. 
This requires manipulation longer than $\SI{32}{\second}$, almost covering the entire sensor surface and including sharp corners. The results are presented in \cref{fig:letters}.

\subsubsection*{Bolt}
\begin{figure}
    \centering
    \includegraphics[width=0.7\linewidth]{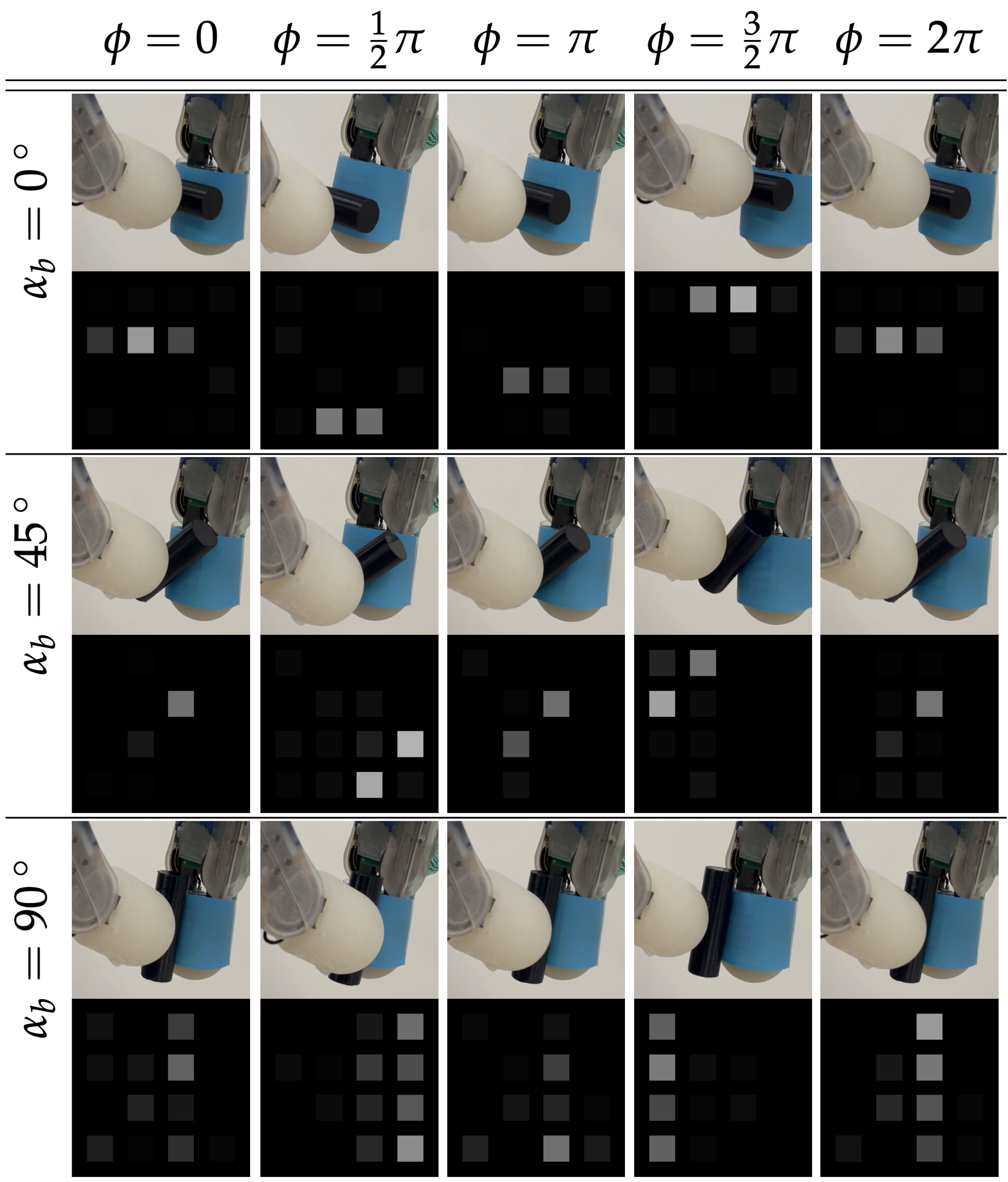}
    \vspace{-0.2cm}
    \caption{Bolt rolling in reality for three different bolt orientations. $\phi$ denotes the phase of the sinusoidal rolling motion, which has a period of $\SI{3}{\second}$.}
    \label{fig:bolt_rolling}
    \vskip -0.5cm
\end{figure}

Moreover, we show the successful sim-to-real transfer of the tactile bolt rolling policy.
For every of the four possible orientations the policy was tested three times.
In all runs, the bolt was rolled in the correct direction.
In $11$ out of the $12$ runs the policy succeeded to run until the timeout after $\SI{24}{\second}$. The failed run dropped the bolt after $\SI{18}{\second}$, as it slipped out the fingers.
Exemplary image sequences of the rolling motion are shown in \cref{fig:bolt_rolling}.

\vspace{-0.04cm}
\section{Conclusion}
We have shown that tactile skin-based fine manipulation with a robotic hand can be learned in simulation and successfully transferred to the real world. For two challenging fine manipulation tasks, we performed detailed simulation experiments and proved the necessity of the tactile feedback. For the marble rolling, the tracking precision was significantly increased from about \SI{5}{\milli\meter} without to less than \SI{1}{\milli\meter} with the tactile skin, reaching clearly sub-taxel precision as the taxel spacing is \SI{4}{\milli\meter}. The task of rolling a bolt along its normal direction even failed completely without the tactile feedback.
Both outcomes prove that the policies depend on the details of the raw tactile signal.
In experiments on the real robotic DLR-Hand~II, the very same in simulation trained policies performed seamlessly and resembled the same qualitatively behavior in both tasks.

To achieve this results, a novel model for the skin was introduced that can be used together with a regular rigid-body physics simulator. To allow a successful sim-to-real transfer, the model was calibrated with real-world data. The calibration process is self-contained and does not require any external sensors. 
To our knowledge, this is the first time that complex fine manipulation of such small objects with robotic fingers was successfully performed on a real robot system -- at least for the case that it was trained purely in simulation.

\scriptsize
\bibliographystyle{IEEEtranN-modified}
\bibliography{IEEEabrv, bibliography}

\begin{thebibliography}{30}
\providecommand{\natexlab}[1]{#1}
\providecommand{\url}[1]{#1}
\csname url@samestyle\endcsname
\providecommand{\newblock}{\relax}
\providecommand{\bibinfo}[2]{#2}
\providecommand{\BIBentrySTDinterwordspacing}{\spaceskip=0pt\relax}
\providecommand{\BIBentryALTinterwordstretchfactor}{4}
\providecommand{\BIBentryALTinterwordspacing}{\spaceskip=\fontdimen2\font plus
\BIBentryALTinterwordstretchfactor\fontdimen3\font minus
  \fontdimen4\font\relax}
\providecommand{\BIBforeignlanguage}[2]{{%
\expandafter\ifx\csname l@#1\endcsname\relax
\typeout{** WARNING: IEEEtranN.bst: No hyphenation pattern has been}%
\typeout{** loaded for the language `#1'. Using the pattern for}%
\typeout{** the default language instead.}%
\else
\language=\csname l@#1\endcsname
\fi
#2}}
\providecommand{\BIBdecl}{\relax}
\BIBdecl

\bibitem[Butterfa\ss et~al.(2001)Butterfa\ss, Grebenstein, Liu, and
  Hirzinger]{Butterfass2001}
J.~Butterfa\ss \emph{et~al.}, ``{DLR-H}and {II}: Next generation of a dextrous
  robot hand,'' in \emph{Proc. IEEE International Conference on Robotics and
  Automation}, 2001.

\bibitem[B{\"a}uml et~al.(2014)B{\"a}uml, Hammer, Wagner, Birbach, Gumpert,
  Zhi, Hillenbrand, Beer, Friedl, and Butterfass]{Bauml.2014}
B.~B{\"a}uml \emph{et~al.}, ``Agile justin: An upgraded member of dlr’s
  family of lightweight and torque controlled humanoids,'' in \emph{Proc. IEEE
  Int. Conf. on Robotics and Automation}, 2014.

\bibitem[Yuan et~al.(2017)Yuan, Dong, and Adelson]{Yuan.2017}
W.~Yuan \emph{et~al.}, ``Gelsight: High-resolution robot tactile sensors for
  estimating geometry and force,'' \emph{Sensors}, vol.~17, no.~12, 2017.

\bibitem[Lambeta et~al.(2020)Lambeta, Chou, Tian, Yang, Maloon, Most, Stroud,
  Santos, Byagowi, Kammerer, et~al.]{Lambeta.2020}
M.~Lambeta \emph{et~al.}, ``Digit: A novel design for a low-cost compact
  high-resolution tactile sensor with application to in-hand manipulation,''
  \emph{IEEE Robotics and Automation Letters}, vol.~5, no.~3, 2020.

\bibitem[Narang et~al.(2021{\natexlab{a}})Narang, Sundaralingam, Van~Wyk,
  Mousavian, and Fox]{Narang.2021b}
Y.~S. Narang \emph{et~al.}, ``Interpreting and predicting tactile signals for
  the syntouch biotac,'' \emph{The International Journal of Robotics Research},
  vol.~40, no. 12-14, 2021.

\bibitem[Narang et~al.(2021{\natexlab{b}})Narang, Sundaralingam, Macklin,
  Mousavian, and Fox]{Narang.2021}
Y.~Narang \emph{et~al.}, ``Sim-to-real for robotic tactile sensing via
  physics-based simulation and learned latent projections,'' in \emph{Proc.
  IEEE Int. Conf. on Robotics and Automation}, 2021.

\bibitem[Moisio et~al.(2013)Moisio, Le{\'o}n, Korkealaakso, and
  Morales]{Moisio.2013}
S.~Moisio \emph{et~al.}, ``Model of tactile sensors using soft contacts and its
  application in robot grasping simulation,'' \emph{Robotics and Autonomous
  Systems}, vol.~61, no.~1, 2013.

\bibitem[Xu et~al.(2023)Xu, Kim, Chen, Garcia, Agrawal, Matusik, and
  Sueda]{JieXu.2023}
J.~Xu \emph{et~al.}, ``Efficient tactile simulation with differentiability for
  robotic manipulation,'' in \emph{Proc. Conf. on Robot Learning}, 2023.

\bibitem[Ding et~al.(2020)Ding, Lepora, and Johns]{Ding.2020}
Z.~Ding \emph{et~al.}, ``Sim-to-real transfer for optical tactile sensing,'' in
  \emph{Proc. IEEE Int. Conf. on Robotics and Automation}, 2020.

\bibitem[Gomes et~al.(2021)Gomes, Paoletti, and Luo]{Gomes.2021}
D.~F. Gomes \emph{et~al.}, ``Generation of gelsight tactile images for sim2real
  learning,'' \emph{IEEE Robotics and Automation Letters}, vol.~6, no.~2, 2021.

\bibitem[Church et~al.(2022)Church, Lloyd, Lepora, et~al.]{AlexChurch.2022}
A.~Church \emph{et~al.}, ``Tactile sim-to-real policy transfer via real-to-sim
  image translation,'' in \emph{Proc. Conf. on Robot Learning}, 2022.

\bibitem[Lin et~al.(2022)Lin, Lloyd, Church, and Lepora]{Lin.2022}
Y.~Lin \emph{et~al.}, ``Tactile gym 2.0: Sim-to-real deep reinforcement
  learning for comparing low-cost high-resolution robot touch,'' \emph{IEEE
  Robotics and Automation Letters}, vol.~7, no.~4, 2022.

\bibitem[Lin et~al.(2023)Lin, Church, Yang, Li, Lloyd, Zhang, and
  Lepora]{Lin.2023}
------, ``Bi-touch: Bimanual tactile manipulation with sim-to-real deep
  reinforcement learning,'' \emph{IEEE Robotics and Automation Letters}, 2023.

\bibitem[Wang et~al.(2022)Wang, Lambeta, Chou, and Calandra]{Wang.2022}
S.~Wang \emph{et~al.}, ``Tacto: A fast, flexible, and open-source simulator for
  high-resolution vision-based tactile sensors,'' \emph{IEEE Robotics and
  Automation Letters}, vol.~7, no.~2, 2022.

\bibitem[Ding et~al.(2021)Ding, Tsai, Lee, and Huang]{Ding.2021}
Z.~Ding \emph{et~al.}, ``Sim-to-real transfer for robotic manipulation with
  tactile sensory,'' in \emph{Proc. IEEE/RSJ Int. Conf. on Intelligent Robots
  and Systems}, 2021.

\bibitem[Yang et~al.(2023)Yang, Huang, Li, Tsai, Lee, Song, and Pan]{Yang.2023}
L.~Yang \emph{et~al.}, ``Tacgnn: Learning tactile-based in-hand manipulation
  with a blind robot using hierarchical graph neural network,'' \emph{IEEE
  Robotics and Automation Letters}, vol.~8, no.~6, 2023.

\bibitem[Habib et~al.(2014)Habib, Ranatunga, Shook, and Popa]{Habib.2014}
A.~Habib \emph{et~al.}, ``Skinsim: A simulation environment for multimodal
  robot skin,'' in \emph{Proc. IEEE Int. Conf. on Automation Science and
  Engineering}.\hskip 1em plus 0.5em minus 0.4em\relax IEEE, 2014.

\bibitem[Tian et~al.(2019)Tian, Ebert, Jayaraman, Mudigonda, Finn, Calandra,
  and Levine]{Tian.2019}
S.~Tian \emph{et~al.}, ``Manipulation by feel: Touch-based control with deep
  predictive models,'' in \emph{Proc. IEEE Int. Conf. on Robotics and
  Automation}, 2019.

\bibitem[Do et~al.(2023)Do, Aumann, Chungyoun, and Kennedy]{Do.2023}
W.~K. Do \emph{et~al.}, ``Inter-finger small object manipulation with densetact
  optical tactile sensor,'' \emph{IEEE Robotics and Automation Letters},
  vol.~9, no.~1, 2023.

\bibitem[Coumans and Bai(2016)]{Coumans.2016}
E.~Coumans and Y.~Bai, ``Pybullet, a python module for physics simulation for
  games, robotics and machine learning,'' 2016.

\bibitem[{Dawson-Haggerty et al.}(2019)]{trimesh}
\BIBentryALTinterwordspacing
{Dawson-Haggerty et al.}, ``trimesh,'' 2019. [Online]. Available:
  \url{https://trimesh.org/}
\BIBentrySTDinterwordspacing

\bibitem[Tenhumberg et~al.(2023)Tenhumberg, Sievers, and
  B{\"a}uml]{Tenhumberg.2023}
J.~Tenhumberg \emph{et~al.}, ``Self-contained and automatic calibration of a
  multi-fingered hand using only pairwise contact measurements,'' in
  \emph{Proc. IEEE-RAS Int. Conf. on Humanoid Robots}, 2023.

\bibitem[Sievers et~al.(2022)Sievers, Pitz, and B{\"a}uml]{Sievers.2022}
L.~Sievers \emph{et~al.}, ``Learning purely tactile in-hand manipulation with a
  torque-controlled hand,'' in \emph{Proc. IEEE Int. Conf. on Robotics and
  Automation}, 2022.

\bibitem[Brockman et~al.(2016)Brockman, Cheung, Pettersson, Schneider,
  Schulman, Tang, and Zaremba]{Brockman.2016}
G.~Brockman \emph{et~al.}, ``Openai gym,'' \emph{arXiv preprint arXiv}, 2016.

\bibitem[Haarnoja et~al.(2018)Haarnoja, Zhou, Abbeel, and
  Levine]{Haarnoja.2018}
T.~Haarnoja \emph{et~al.}, ``Soft actor-critic: Off-policy maximum entropy deep
  reinforcement learning with a stochastic actor,'' in \emph{Int. Conf. on
  Machine Learning}, 2018.

\bibitem[Raffin et~al.(2021)Raffin, Hill, Gleave, Kanervisto, Ernestus, and
  Dormann]{Raffin.2021}
A.~Raffin \emph{et~al.}, ``Stable-baselines3: Reliable reinforcement learning
  implementations,'' \emph{The Journal of Machine Learning Research}, vol.~22,
  no.~1, 2021.

\bibitem[Kingma and Ba(2015)]{Kingma.22.12.2014}
D.~Kingma and J.~Ba, ``Adam: A method for stochastic optimization,'' in
  \emph{Int. Conf. on Learning Representations}, 2015.

\bibitem[Andrychowicz et~al.(2020)Andrychowicz, Baker, Chociej, Jozefowicz,
  McGrew, Pachocki, Petron, Plappert, Powell, Ray, et~al.]{Andrychowicz.2020}
M.~Andrychowicz \emph{et~al.}, ``Learning dexterous in-hand manipulation,''
  \emph{The International Journal of Robotics Research}, vol.~39, no.~1, 2020.

\bibitem[Akkaya et~al.(2019)Akkaya, Andrychowicz, Chociej, Litwin, McGrew,
  Petron, Paino, Plappert, Powell, Ribas, et~al.]{OpenAI.2019}
I.~Akkaya \emph{et~al.}, ``Solving rubik's cube with a robot hand,''
  \emph{arXiv preprint}, 2019.

\bibitem[Pitz et~al.(2023)Pitz, R{\"o}stel, Sievers, and B{\"a}uml]{Pitz.2023}
J.~Pitz \emph{et~al.}, ``Dextrous tactile in-hand manipulation using a modular
  reinforcement learning architecture,'' in \emph{Proc. IEEE Int. Conf. on
  Robotics and Automation}, 2023.

\end{thebibliography}

\end{document}